\documentclass{article}

\pdfoutput=1
\PassOptionsToPackage{numbers, compress}{natbib}

\usepackage[final]{neurips_2023}

\usepackage[utf8]{inputenc} 
\usepackage[T1]{fontenc}    
\usepackage{hyperref}       
\usepackage{url}            
\usepackage{booktabs}       
\usepackage{amsfonts}       
\usepackage{nicefrac}       
\usepackage{microtype}      
\usepackage{xcolor}         

\usepackage{color, soul}
\usepackage{tabularray}
\definecolor{Romantic}{rgb}{1,0.827,0.713}
\definecolor{CornflowerLilac}{rgb}{1,0.666,0.643}
\definecolor{Padua}{rgb}{0.658,0.898,0.811}
\definecolor{ChromeWhite}{rgb}{0.862,0.925,0.756}
\definecolor{LinkWater}{rgb}{0.807,0.85,0.956}
\definecolor{Milan}{rgb}{1,0.996,0.639}
\usepackage{rotating}
\usepackage{enumitem}
\usepackage{tikz}
\usepackage{float}
\hypersetup{breaklinks=true}

\usepackage{cleveref}
\crefname{section}{\S}{\S}
\crefname{Section}{\S}{\S}
\crefformat{section}{§#2#1#3}
\crefname{table}{Tab.}{Tab.}
\crefname{appendix_table}{Tab.}{Tab.}
\crefname{Table}{Tab.}{Tab.}
\crefname{Figure}{Fig.}{Fig.}
\crefname{figure}{Fig.}{Fig.}
\crefname{appendix}{Appendix}{Appendix}
\crefname{chapter}{Chapter}{Chapter}

\title{The Empty Signifier Problem:\\ Towards Clearer Paradigms for Operationalising ``Alignment'' in Large Language Models}

\author{%
  Hannah Rose Kirk$^{1}$
  \And
  Bertie Vidgen$^{1,2}$
    \And
  Paul Röttger$^{3}$
  \And
  Scott A. Hale$^{1,2,4}$
  \And
  \normalfont{${^1}$University of Oxford, ${^2}$The Alan Turing Institute, ${^3}$Bocconi University, ${^4}$Meedan}
}

\begin{document}

\maketitle

\begin{abstract}
In this paper, we address the concept of ``alignment'' in large language models (LLMs) through the lens of post-structuralist socio-political theory, specifically examining its parallels to empty signifiers. To establish a shared vocabulary around how abstract concepts of alignment are operationalised in empirical datasets, we propose a framework that demarcates: 1) which dimensions of model behaviour are considered important, then 2) how meanings and definitions are ascribed to these dimensions, and by whom. We situate existing empirical literature and provide guidance on deciding which paradigm to follow. Through this framework, we aim to foster a culture of transparency and critical evaluation, aiding the community in navigating the complexities of aligning LLMs with human populations.
\end{abstract}
\section{Introduction}
In post-structuralist socio-political theory, empty signifiers are terms or symbols characterised by their absence of fixed referents \cite{levi-straussIntroduction1987, laclauEmancipation1996}. Acting as discursive placeholders, these vague and abstract terms are infused with meaning by different individuals or groups, who can exploit the ambiguity for their particular negotiations of universally-known, but not universally-defined, concepts. In his work \textit{Emancipation}, Laclau distinguishes \textit{empty signifiers} from the related concept of \textit{floating signifiers} \cite{laclauEmancipation1996}. A floating signifier absorbs meaning, allowing the signified concepts to be interpreted fluidly and contextually. In contrast, as Laclau argues, empty signifiers have a stronger power implication---they are terms devoid of meaning precisely because different social and political groups ascribing particularistic meaning is what maintains and drives hegemonic order. In this political context, empty signifiers serve as vehicles for ambiguous or notionally ``universal'' agreeable pursuits, rallying disparate individuals around abstract ideas without ever offering a concrete conceptual anchor.

The notion of ``alignment'' in large language models and other AI systems has attracted unprecedented attention in the past year, from researchers, developers, policymakers and citizens alike. Akin to empty signifiers, the term ``alignment'' serves as a rhetorical placeholder for an aspirational conceptualisation of relations between humans and machines, which is fairly unobjectionable in principle, but lacks a shared definition or goal to translate in practice \cite{gabrielArtificial2020a}.\footnote{Some take it to mean aligning AI with our \textit{standards} \cite{turingLecture1947}, \textit{wants} \cite{kentonAlignment2021} or \textit{motives} \cite{christianoClarifying2021}; \textit{revealed, stated} or \textit{idealised preferences} \cite{gabrielArtificial2020a}; \textit{communication norms} \cite{kasirzadehConversation2022}; \textit{intents}, or \textit{expectations} \cite{ouyangTraining2022, leikeScalable2018b} and \textit{goals} \cite{irvingAI2018}; other focus specially on \textit{value} alignment \cite{russellHuman2019, kosterHumancentred2022} or \textit{social} alignment \cite{liuTraining2023}; finally, there are some who interpret it in the limit as avoiding harm and suffering \cite{pengReducing2020} or even mitigating far-future notions of `x-risk'.} Statements---like ``ensure that powerful AI is properly aligned with human values'' \cite[p.2,][]{russellHuman2019} or ``text-based assistant that is aligned with human values'' \cite[p.1,][]{askellGeneral2021} or ``the behaviour of AI agents needs to be aligned with what humans want'' \cite[p.1,][]{kentonAlignment2021} or how to generate text that is ``in accordance with some shared human values'' \cite[p.243,][]{liuAligning2022}---appeal to vague and fuzzy ideals, yet do not confront the complexities of what these statements imply nor critically reflect on whose meanings, and which power structures, are encoded in reality. 

Empty signifiers, and by analogy these wide-sweeping conceptualisations of alignment, are only sustainable in so far as the abstract can remain abstract. Yet, to \textit{empirically} align language models, that is to tangibly steer them towards certain behaviours and away from others, we require some form of measurable signal or data. When designing data collection protocols, writing annotator guidelines, or hiring annotators, abstract notions of alignment must be calcified into \textit{which preferences, values, or behaviours are important} and \textit{how to measure them}. Just like Laclau's empty signifiers, this question cannot be tangibly answered without the tainting of identity, politics and power. In practice, the question of ``which'' becomes a question of ``whose'' \cite{gabrielArtificial2020a}---both in terms of who decides the properties of a model that is ``aligned'', and who actually interprets these concepts for data labelling or curation, and feedback provision. As \citet{ouyangTraining2022} state: ``one of the biggest open questions is how to design an alignment process that is transparent'' (p.19). In this article, we confront the practical complexities of operationalising abstract notions of alignment into observable training or evaluation signals, in order to encourage greater transparency and shared understanding.

There is a large growing literature that collects empirical data recording human feedback, demonstration and instruction for steering language model behaviour under the motivation of ``alignment''. This literature has transitioned from early work that adopted abstract and general notions of human preference, like ``goodness'' or ``quality'' \cite{stiennonLearning2020, zieglerFineTuning2019} towards collecting more fine-grained, disaggregated and rich feedback \cite{scheurerTraining2022, wuFineGrained2023, chengEveryone2023}, or being explicit about the targeted traits and behaviours \cite{baiTraining2022, askellGeneral2021, thoppilanLaMDA2022, glaeseImproving2022}. Presenting the literature in detail is outside the remit of this paper but we draw closely on the body of work recently surveyed by \citet{kirkPast2023} and \citet{wangAligning2023}. When scrutinising the general discourse around human feedback learning---be it red-teaming \cite{ganguliRed2022}, reinforcement learning with human feedback \cite{ouyangTraining2022, nakanoWebGPT2021, glaeseImproving2022, baiTraining2022}, preference pre-training \cite{korbakPretraining2023}, supervised fine-tuning \cite{zhouLIMA2023} or direct preference optimisation \cite{rafailovDirect2023a}---there is a lack of shared terminology for articulating what alignment actually means in a given empirical context, what it tangibly achieves, and for whom. Confronting the practical difficulties of converting subjective concepts into labelled or categorised data is not a new problem, and inspiration can be drawn from neighbouring fields. Interpreting annotator differences as signal not noise \cite{inelCrowdTruth2014a, aroyoTruth2015, nie2020can}, modelling disagreements \cite{davani2022dealing} or annotator artefacts \cite{gururanganAnnotation2018, sapAnnotators2022c}, releasing detailed documentation \cite{prabhakaranReleasing2021b, benderData2018b}, and doing away with majority vote in favour of more nuanced voting or aggregation systems \cite{gordon2022jury} are established practices in other areas of natural language processing, as well as computer vision \cite{sharmanskaAmbiguity2016, schmarjeOne2022, parrishPicture2023, collinsEliciting2022}.

In this paper, we extend a framework for annotating subjective NLP datasets from \citet{rottgerTwo2022} to the creation of datasets for LLM alignment. The original framework introduces two data annotation paradigms for facilitating different end goals.
The \textit{prescriptive paradigm} discourages annotator subjectivity by providing detailed guidelines, to encode a single set of beliefs in the data.
The \textit{descriptive paradigm}, on the other hand, encourages subjectivity, with the goal of capturing a diversity of beliefs.
\citet{rottgerTwo2022} give the example of \textit{prescriptive} enforcement of hate speech policies on large online platforms, compared to \textit{descriptive} analyses of different perceptions of online hate.
By introducing the two paradigms, they hope to enable better documentation and more clarity about the intended uses of different datasets. In the domain of collecting data for LLM alignment, we echo this call for a clearer articulation of objectives. For alignment, it is just as necessary to ask ``how to operationalise and communicate a certain concept''.
However, for alignment, we also need to ask ``how to decide which concepts are relevant'' in the first place. Our proposed framework centers these two decision points: 
 \begin{enumerate}[noitemsep]
     \item Identifying the \textbf{dimensions} that are included as in scope for the alignment dataset, which can either be \textit{Broad} (general goals e.g., ``outputs that people prefer'', ``good outputs'') or \textit{Specific} (named traits or behaviours e.g., ``honestly'', ``informativeness'', ``harmlessness'')
     \item Determining the interpretative authority and invariability of the \textbf{definitions} of these dimensions, which can be \textit{Prescriptive} (clear detailed definitions of a single belief) or \textit{Descriptive} (subjective interpretation and many beliefs) \cite{rottger2022two}.
 \end{enumerate}

Our motivation for introducing this new conceptual framework is to foster a culture of transparency and critical consideration within the empirical alignment research community. To demonstrate its practical utility, we communicate an \textit{ex-post} mapping of existing literature onto the framework's axes; but the framework is intended as a tool for clarifying and communicating \textit{ex-ante} intents of the dataset designers and model builders. Our proposed mapping inevitably suffers from incompleteness in ways that alignment datasets can be categorised; as well as idiosyncrasies of individual researchers or practitioners adopting the role of cartographer. Nonetheless, even if the boundaries and characterisations are imperfect, our aim is to equip people with a language to specify their intents. As more and more people across the world use LLMs on a regular basis, the boundaries of what it means for a model to be aligned, to be helpful, honest or harmless, become increasingly intractable and fuzzy. We hope our framework will assist in transmuting the empty signifier of ``alignment'' into actionable constructs, empirical signals and sampling strategies, that can be communicated, criticised, and re-conceptualised in a common language.
\section{Paradigms for Operationalising Alignment}

\begin{figure}
    \centering
    \resizebox{0.585\linewidth}{!}{
    \begin{tikzpicture}
        \path[use as bounding box] (-5.25,-5.25) rectangle (5.25,5.25); 
        \fill[Padua, opacity=0.4] (0,0) rectangle (5,5);   
        \fill[Romantic, opacity=0.4] (-5,0) rectangle (0,5); 
        \fill[ChromeWhite, opacity=0.4] (-5,-5) rectangle (0,0); 
        \fill[CornflowerLilac, opacity=0.4] (0,-5) rectangle (5,0); 

        \draw[thick,->] (-5,0) -- (5,0) node[pos=0.75, align=center, below] {\textbf{Selectivity of DIMENSIONS}} node[above, right] at (5,0) {\textit{\large{SPECIFIC}}};
        \draw[thick,->] (0,-5) -- (0,5) node[pos=0.75, above, rotate=90] {\textbf{Invariability of DEFINITIONS}} node[align=center] at (0,5.25) {\textit{\large{PRESCRIPTIVE}}};

        \node[above, left] at (-5,0) {\textit{\large{BROAD}}};
        \node[align=center] at (0,-5.25) {\textit{\large{DESCRIPTIVE}}};


         \node[draw, rounded corners, fill=gray!20, opacity=0.7, text opacity=1, align=center, inner sep=2pt] at (2,1) {Google LaMDA \\ human feedback \cite{thoppilanLaMDA2022}};
          \node[draw, rounded corners, fill=gray!20, opacity=0.7, text opacity=1, align=center, inner sep=2pt] at (2.9,2.5) {OpenAI WebGPT \\ human feedback \cite{nakanoWebGPT2021}};
        \node[draw, rounded corners, fill=gray!20, opacity=0.7, text opacity=1, align=center, inner sep=2pt] at (3.5,4) {OpenAI ChatGPT\\guidelines for \\reviewers \cite{openaiSnapshot2023}};
        \node[draw, rounded corners, fill=gray!20, opacity=0.7, text opacity=1, align=center, inner sep=2pt] at (1.1,4) {Baidu\\open\\domain\\chatbot \cite{luBoosting2022}};
        
        \node[draw, fill=gray!20, opacity=0.7, text opacity=1, align=center, rounded corners, inner sep=2pt] at (-3,3.5) {OpenAI summarisation \\ with human feedback \\ \cite{zieglerFineTuning2019, stiennonLearning2020} \\(and long-form book \\ summarisation \cite{wuRecursively2021})};
        \node[draw, fill=gray!20, opacity=0.7, text opacity=1, align=center, rounded corners, inner sep=2pt] at (-2.5,1.7) {OpenAI \\InstructGPT \cite{ouyangTraining2022}};
        \node[draw, fill=gray!20, opacity=0.7, text opacity=1, align=center, rounded corners, inner sep=2pt] at (-2,0.5) {DeepMind \\GopherCite \cite{menickTeaching2022}};
        \node[draw, fill=gray!20, opacity=0.7, text opacity=1, align=center, rounded corners, inner sep=2pt] at (-3,-3) {Stanford Human\\ Preferences dataset \cite{stanfordnlpStanford2023}};
        \node[draw, fill=gray!20, opacity=0.7, text opacity=1, align=center, rounded corners, inner sep=2pt] at (-3.5,-4.2) {LM-SYS \\ Chat-1M \cite{zhengLMSYSChat1M2023}};
        \node[draw, fill=gray!20, opacity=0.7, text opacity=1, align=center, rounded corners, inner sep=2pt] at (-2,-1) {DeepMind\\ diverse consensus \\ fine-tuning \cite{bakkerFinetuning2022}};
        
        \node[draw, fill=gray!20, opacity=0.7, text opacity=1, align=center, rounded corners, inner sep=2pt] at (2,-1.2) {OpenAI GPT-4(V) \\ red-teaming \cite{openaiGPT4V2023}};
        \node[draw, fill=gray!20, opacity=0.7, text opacity=1, align=center, rounded corners, inner sep=2pt] at (3.5,-2.7) {Anthropic\\ harmlessness \cite{ganguliRed2022}};
        \node[draw, fill=gray!20, opacity=0.7, text opacity=1, align=center, rounded corners, inner sep=2pt] at (3.75,-4.2) {Anthropic\\ helpfulness \& \\ honesty \cite{askellGeneral2021, baiTraining2022}};
        \node[draw, fill=gray!20, opacity=0.7, text opacity=1, align=center, rounded corners, inner sep=2pt] at (1.5,-4) {Anthropic\\constitution\\writing \\\cite{baiConstitutional2022}};
    \end{tikzpicture}
    }
    \caption{\small A mapping of existing datasets and empirical literature to our framework. See \cref{tab:2by2} for full descriptions and quotations.}
    \label{fig:quadrants}
\end{figure}
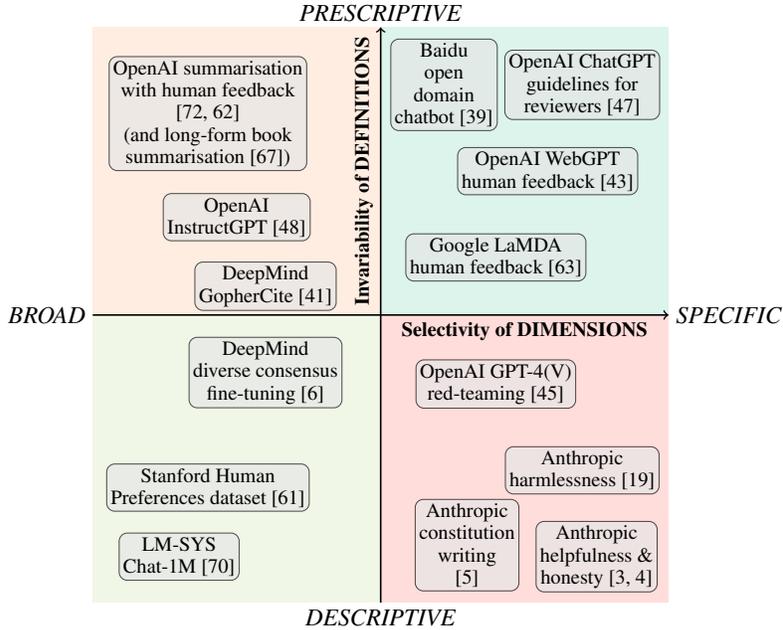

The process of creating alignment data encompasses two core decisions: 1) selecting dimensions to be included, and 2) deciding how dimensions are defined. We think these decisions are most relevant to what, how and whose behaviours get encoded into LLMs during alignment-tuning;\footnote{There are established norms and terminologies for some axes of dataset curation and design in shared documentation standards like Data Statements \cite{bender2018data} or Datasheets \cite{gebru2021datasheets}. We intentionally do not focus on these aspects and instead on what we deem crucial for understanding alignment's interpretative scope and subjectivity.} but many other decision points exist, like the weightings of dimensions, aggregation functions, or presentation and order of decisions in interfaces.\footnote{For example, \citet{nakanoWebGPT2021} provide ``criteria in descending order of priority'' for raters making final decisions. The criteria include ``Whether or not the answer contains unsupported information'', and ``How much irrelevant information there is in the answer (This can be higher priority in extreme cases.)'' (p.18).} There are many ways of collecting empirical signals for alignment \cite{kirkPast2023} and our framework is applicable and adaptable to many forms: comparisons between outputs (pick the more \textit{<helpful>} output from this set) \cite{zieglerFineTuning2019, askellGeneral2021, stiennonLearning2020, baiTraining2022, ouyangTraining2022}; demonstrations (write a \textit{<helpful>} answer to this query) \cite{nakanoWebGPT2021}; rewrites or edit chains (rewrite the output to be more \textit{<helpful>}) \cite{liuSecond2023}; or natural language feedback (explain why the output is/is not \textit{<helpful>}) \cite{scheurerTraining2022}. In this section, we outline our 2x2 framework, discuss how a researcher or developer locates themselves, and demonstrate how existing literature maps onto each of the four quadrants (see \cref{fig:quadrants} and \cref{tab:2by2}).

\subsection{The 2x2 Framework}
\paragraph{Dimensions} Any decision to measure a particular attribute of model behaviour (e.g., ``Honesty'') already bounds the alignment input-output space. The first level of our framework contends with the one or more \textit{dimensions} that are considered `in-scope' for the empirical alignment dataset. On one end of the spectrum is the \textit{Broad} paradigm, where some high-level concept is targeted, without fine-grained detail or sub-behaviours. For example, this could be collecting data on ``which output do you prefer'', ``what output is better'', ``how good is this output'', or ``rate the quality of each output'' \cite{zieglerFineTuning2019, stiennonLearning2020}. Examining ELO ratings between model outputs that people prefer is a canonical example of the \textit{Broad} paradigm \cite{zhengLMSYSChat1M2023}.\footnote{For example, an interface like \href{https://chat.lmsys.org/}{chat.lmsys} which only includes ``A(B) is better'', ''tie'' or ``both are bad''.} On the other end of the spectrum is the \textit{Specific} paradigm, where technology providers and dataset creators centre their empirical efforts on named attributes or targeted traits. For example, they decide \textit{a priori} that ``honesty, harmlessness, and helpfulness'' \cite{askellGeneral2021, baiTraining2022}, or ``informativeness'' \cite{nakanoWebGPT2021} and ``safety'' \cite{thoppilanLaMDA2022}, are key traits to get right for an aligned model. Note that the definitions of these specified dimensions can still be left open to interpretation.

\paragraph{Definitions} Adopting \citet{rottger2022two}'s terminology, the conceptual treatment of definitions can either be \textit{Prescriptive} or \textit{Descriptive}. Under the \textit{Prescriptive} paradigm, technology providers or dataset creators write detailed guidelines or training tasks, thus pinning down a narrow interpretation and asking annotators to abide by this interpretation as far as possible \cite{nakanoWebGPT2021, glaeseImproving2022, luBoosting2022, thoppilanLaMDA2022, openaiSnapshot2023}. Under this paradigm, practitioners measure and try to maximise inter-annotator agreement or calculate majority votes, where conceptually any deviations from agreement represents noise, misunderstanding of the guidelines or poor quality annotation. In contrast, the \textit{Descriptive} paradigm avoids defining the meaning signified by different dimension terms, instead allowing human raters to subjectively inscribe their own contextual meanings \cite{ganguliRed2022, baiTraining2022}. Under this paradigm, interpersonal disagreements are not sought to be minimised; in fact, such differences may be the exact object of interest for the study.

\subsection{Questioning the End Goal}
Similar to \citet{rottgerTwo2022}, we suggest any researcher or developer carefully considers the purpose of their empirical alignment dataset in order to decide how it is scoped and collected.

\paragraph{Broad or Specific Dimensions?} Three questions guide this choice. First, \textbf{how well can I define my aim?} \textit{Broad} dimensions can identify a wide landscape of important behaviours (that may include `unknown unknowns'), without biasing data with \textit{a priori} expectations. This adopts a positivist epistemology---that the `reality' or legitimacy of phenomena emerges from empirical measurement. Conversely, a \textit{Specific} stance more reliably encodes particular dimensions considered intrinsically or instrumentally important (for profit, political, reputational or social motives).\footnote{There are trade-offs to consider. The dimensions in a \textit{Specific} stance may be at odds with or deprioritise other critical dimensions \cite{askellGeneral2021, baiTraining2022}. The flexibility of the \textit{Broad} stance loses information on \textit{why} people prefer one output over another or potentially encourages raters to take shortcuts on artefacts such as output length \cite{wuRecursively2021}.} Second, \textbf{how well can I estimate applications of a model tuned on my data?} The narrower the usecase the easier to specify dimensions: a model used for only fiction summarisation or creative writing requires a different and narrower set of traits than a model for information-seeking dialogue \cite{askellGeneral2021} or a general purpose conversational agent \cite{askellGeneral2021, baiTraining2022}.
Lastly, \textbf{how well can I assess what's needed to accomplish the task?} Task complexity influences the ease of specifying dimensions---inverse reinforcement learning was precisely conceived for when oversight is challenging \cite{bowmanMeasuring2022}; accordingly, if dimensions cannot be reliably outlined or validated, we recommend a \textit{Broad} approach.

\begin{table}
\footnotesize
\caption{Empirical literature and existing datasets situated in our framework. We present examples for each of the four quadrants to demonstrate the practical differences between paradigms.}
\label{tab:2by2}
\centering
\resizebox{\textwidth}{!}{  
\begin{tblr}{
  colspec = {c  p{2cm} | p{6.5cm} | p{6.5cm}}, 
  cells = {halign=c, valign=m},
  cell{1}{3} = {c=2}{},
  cell{2}{3} = {ChromeWhite},
  cell{2}{4} = {Padua},
  cell{3}{1} = {r=2}{},
  cell{3}{2} = {Romantic},
  cell{4}{2} = {CornflowerLilac},
  cell{5}{3} = {valign=t}, 
  cell{5}{4} = {valign=t}, 
  cell{7}{3} = {valign=t}, 
  cell{7}{4} = {valign=t}  
}
 &  & \Large{\textbf{I. DIMENSIONS}} & \\
 &  & {\large{\textbf{Broad}}\\\normalsize{Leave open wide scope for any or many dimensions of model behaviour.}} & {\large{\textbf{Specific}}\\\normalsize{Focus explicitly on one or more \textit{named} dimensions of model behaviour.}}\\
 
\begin{sideways}\Large{\textbf{II. DEFINITIONS}}\end{sideways} & {\large{\textbf{Prescriptive}} \\\normalsize{Provide a single meaning of \textit{dimensions} via detailed guidelines and definitions.}}
& \begin{itemize}[leftmargin=*, noitemsep]
    \item \citet{stiennonLearning2020} and \citet{wuRecursively2021} very generally seek ``good'' summaries, asking annotators ``which summary is best''. Note that they do condition on length: ``how good is this summary, given it is X words long?'', but this still represents a \textit{broad} position. They then however describe in detail what properties a ``good'' summary has and even prescribe sub-dimensions like coherence, purpose and style.  For example, ``roughly speaking, a good summary is a shorter piece of text that has the same essence of the original – tries to accomplish the same purpose and conveys the same information as the original post'' \cite[p.21,][]{stiennonLearning2020}
   \item  \citet{wuRecursively2021} is a bit of an edge case, because they externally focus on ``quality'' of the summary but in reality, they very precisely define subconcepts of coverage, coherence and amount of abstraction. Their guidelines even include phrases like ``Present tense should be preferred'' (see p.23 for the full detailed guidelines and definitions).
    \item \citet{zieglerFineTuning2019} simply ask which outputs are preferred but give detailed instructions for labellers and provide example comparisons labelled by the authors.
    \item \citet{menickTeaching2022} also target general concepts of ``good'' or ``bad'' answers but then in guidelines, make statements like ``helpful answers are better'' and ``answers which make you read less are better'' (p.29, weakly prescriptive).
\end{itemize}
& \begin{itemize}[leftmargin=*, noitemsep]
    \item \citet{openaiSnapshot2023} focus on reducing political bias and improving handling of controversy in ChatGPT. Intended behaviours are defined in precise and detailed guidelines given to human reviewers, containing edge cases and exemplars.
    \item \citet{luBoosting2022} give precise definitions and detailed guidelines on how to interpret coherence, informativeness, safety and engagingness. For evaluation, they take majority vote across three annotators and report interannotator agreement (which is low suggesting remaining subjective scope.)
    \item \citet{thoppilanLaMDA2022} define fairly clear targets of informativeness, safety and quality, then give detailed guidelines and definitions for how to interpret these concepts. They make it explicit that a single organisational perspective is sought, where \textit{Safety} is defined according to ``objectives derived from Google's AI Principles'' (p.5). However, there are still some elements of slight subjectivity because they appeal to annotators to ``use their commonsense'' (p.34).
    \item \citet{nakanoWebGPT2021} specify helpfulness (informativeness) as the key required trait in their WebGPT model, designed for information-seeking dialogue. They provide contractors with a video and detailed instructions ``to enable more
interpretable and consistent comparisons'' (p.6), ``to minimize label noise'' (p.17), and ``to make comparisons as unambiguous as possible'' (p.17).
\end{itemize}\\ \cline{3-5}

 & {\large{\textbf{Descriptive}}\\\normalsize{Allow and encourage subjective interpretation of \textit{dimensions} by providing no definitions.}}
& \begin{itemize}[leftmargin=*, noitemsep]
    \item \citet{zhengLMSYSChat1M2023} collect ELO scores from internet users rating different LLMs and collate them into a large-scale dataset. The interface only asks which model output is ``better'', if both are ``bad'' or if there is a tie. It does not specify which dimensions could (or should) contribute to ``better'' nor define what ``better'' means.
    \item The \citet{stanfordnlpStanford2023} Human Preferences (SHP) dataset is sourced from upvoting behaviours on different subredddits. While they specify in the Data Card that upvotes corresponds to ``helpfulness'', we consider this an example of broad-descriptive because in reality, there are no specifications or definitions of dimensions that a Reddit user looks for to upvote one post over another---it is a very broad and descriptive signal of preference.
    \item \citet{bakkerFinetuning2022} could also be considered broad-descriptive because they explicitly seek to collect divergent opinions on a wide range of moral and political topics, and provide little prescription over what opinions individuals can hold. However, they do then target agreement and quality as desirable properties of an opinion consensus summary (adding in some specific dimensions). 
    \item \citet{liuAligning2022}, in their evaluations using human annotators, ask ``How much do you agree that the generated text is aligned with the human value: morality/deontology/non-toxicity?'' (p.248). Similarly broad, \citet{liuSecond2023} ask ``To what extent does the edited response improve the original response in terms of alignment with human values?'' (p.6). This evaluation setting represents the most broad-descriptive statement we can imagine because they ask directly about the meta-goal (and empty signifier) of ``alignment'' with no dimensions and no definitions.
\end{itemize}
& \begin{itemize}[leftmargin=*, noitemsep]
    \item \citet{openaiGPT4V2023} seek to identify unsafe behaviours by red-teaming GPT-4(V). They do not \textit{a priori} define all the different instantiations of ``unsafe'' according to OpenAI's principles, and instead hire experts to locate and interpret areas of risk.
    \item \citet{ganguliRed2022} specify the target behaviour (harm) but offer no clear definitions. They say: ``We do not define what ``harmful'' means, as this is a complex and subjective concept; instead, we rely on the red team to make their own determinations via a pairwise preference choice'' (p.4). Note that they do still measure and report inter-annotator agreement between raters and authors (p.8-9).
    \item \citet{askellGeneral2021} and \citet{baiTraining2022} stipulate that a value-aligned model is one that is honest, harmless and helpful but do not prescribe their meaning. \citet{baiTraining2022} say ``we certainly believe that honesty is a crucial goal for AI alignment'' (p.4), but also ``[o]ur goal is not to define or prescribe what `helpful' and `harmless' mean but to evaluate the effectiveness of our training techniques, so for the most part we simply let our crowdworkers interpret these concepts as they see fit.'' (p.4).
    \item \citet{baiConstitutional2022} apply a specific-descriptive framing for writing the constitutions in their constitutional AI framework. Some constitutions include statements like ``Identify all ways in which the assistant’s last response is harmful, unethical, or socially biased'' (p.22), without defining these concepts (despite the possibility for substantial variation across ethical frameworks and different societal structures or hierarchies by community, culture or country). Some of the constitutions are more prescriptive in defining sub-concepts like \textit{harmful}, for example referencing racist, sexist, toxic, illegal, violent, or unethical behavior.
\end{itemize}\\
\end{tblr}
}
\vspace{-3em}
\end{table}

\paragraph{Descriptive or Prescriptive Definitions?} The key question here is \textbf{do I want to encode a specific belief or capture a diversity of beliefs?} \cite{rottger2022two}. There is a wave of new literature seeking to measure sociocultural and interpersonal variation in human perceptions of language model outputs \cite{xueReinforcement2023, chengEveryone2023, aroyoDICES2023}, as well as assess whether language models themselves reflect diverse opinions \cite{durmusMeasuring2023,hallerOpinionGPT2023} or generate consensus \cite{bakkerFinetuning2022}. Understanding and/or incorporating diversity in alignment perspectives requires the \textit{Descriptive} paradigm, especially if targeting personalisation or customisation \cite{kirk2023personalisation, salemiLaMP2023, bhattLearning2023}. Encoding one belief requires the \textit{Prescriptive} paradigm to communicate concepts consistently via training tasks or detailed guidelines; yet even with these interventions, disagreement cannot be fully eliminated \cite{glaeseImproving2022, ouyangTraining2022, stiennonLearning2020}. Nevertheless, practitioners should clearly communicate whether disagreement between people is the signal they seek to study or the noise they work to eliminate \cite{rottger2022two}.\footnote{Statements from \citet{zieglerFineTuning2019} demonstrate conflicting statements between paradigms. In the same page (p.12), they say ``Evaluation of a summary is both subjective and multidimensional'' (\textit{Descriptive}), and ``One could hope to cope with such `noise' by simply getting more labels and averaging them but this does not resolve all the practical difficulties with ambiguity'' (\textit{Prescriptive}).}
\section{Discussion}
\paragraph{Appreciating that everyone knows what a horse is} Our framework relies on an interplay between dimensions and definitions, but it is not practical to treat all dimensions equally. \textit{Noew Ateny}, a Polish dictionary published in 1745 amusingly includes the definition: ``Horse: Everyone can see what a horse is''. In a similar vein, it is apparent that evaluating dimensions like ``length'' of an output face much less debate than societally-subjective dimensions like ``harmlessness'' or concepts that can be ascribed many different sub-meanings like ``good''. This implies that the levels of our framework inherently interact with one another because dimensions carry with them an intrinsic or expected level of objectivity, which in turn conditions the necessity or impact of providing a definition.

\paragraph{Disputing alignment at the margins, not the extremes}
We expect a lesser degree of disagreement on important dimensions and their definitions at the extremes. While values, morals and ethics vary considerably across individuals, cultures and time \cite{fischerPersonalityValuesCulture2017}, there are universally-agreed bounds on the range of variations, evident in human rights declarations. These bounds address fundamental basic rights, like the right to life, but remain vague on their practical application in contentious issues like abortion. Similarly for empirical alignment efforts, distinctions between \textit{specific-prescriptive} and \textit{broad-descriptive} positions converge at the extremes: it is a sensible assumption that annotators likely internalise that LLMs shouldn't seek to harm life or property, even if this behaviour is unspecified. That said, boundaries may need to placed on the acceptable range or limit of inclusions and interpretations \cite{kirk2023personalisation}, and failing to do so may result in degenerative outcomes, as seen with the Tay Bot incident \cite{kimMimetic2018}. Determining these boundaries is a complex normative issue but a partial solution comes from deciding who forms the sample or pool that provides the alignment signals.

\paragraph{Communicating universals versus particulars} Adopting a particular position does not necessarily convey a given philosophy but may add colour to underlying thinking. Placing very weak restrictions on how language models should behave (\textit{broad-descriptive}) is congruous to a cultural relativist position, where individual identity and lived experience condition meaning. Adopting a \textit{specific-prescriptive} stance can underpin a multitude of belief systems. People making locally-bounded assumptions on how models should behave can acknowledge diverse interpretations yet encode specific ``designer preferences'' or organisational priorities into models  \cite{stiennonLearning2020, thoppilanLaMDA2022, openaiHow2023}. Those making more global and unbounded assumptions may buy into shared interpretations and universalities across different cultures, time periods and peoples. In any approach, it is important to communicate the role that identity and positionality has on scoping, prescription and interpretation, and whether representativeness of the sample conditions the value of the empirical signal. 
There is growing acknowledgement of how identity conditions reward \cite{chengEveryone2023} or risk \cite{aroyoDICES2023, ganguliRed2022}; yet some widely-used datasets \cite[e.g.,][]{baiTraining2022} make no provisions on annotator identity despite being \textit{descriptive} in scope. In the absence of clear communication, there is a risk of conflating the particular and the universal, as Judith Butler comments: ``the universalization of the particular seeks to elevate a specific content to a global condition, making an empire of its local meaning'' \cite[p.31,][]{butlerContingency2000} 

\paragraph{Acknowledging power dynamics and hegemonies} Our framework promotes clear articulation of stakeholder influence in empirical alignment efforts, aiding the understanding of power hierarchies. In the \textit{specific-prescriptive} stance, organisations or model developers impose top-down restrictions. Here, honest communication around how decisions were made and why includes both explaining inclusions---which \citet{ouyangTraining2022} does very well in their attribution of researchers, labellers and OpenAI's role in shaping encoded preferences---and being upfront about the possibility of exclusions---which \citet{thoppilanLaMDA2022} exemplify in discussing sociocultural and geographical blindspots in their definition of safety. However, even in absence of clear prescription, there is still a risk for hegemonic reinforcement, which could be even more dangerous when disguised. While the \textit{broad-descriptive} position might seem most accepting of human variation and sociocultural difference as a ``bottom-up'' grassroots approach, it is often a non-representative group who shoulder the responsibility for steering LLM behaviours---whether few US-based crowdworkers \cite[e.g., see][]{baiTraining2022, nakanoWebGPT2021} or many interested netizens \cite{zhengLMSYSChat1M2023}. While our framework does not prescribe who should have a voice in LLM development, it does encourage greater admission of individual, community and organisational involvement. Without these contextual bounds, there is a risk of moral absolutism as a form of digital colonialism, where the values, morals or priorities of US-based technology providers and crowdworkers are imposed on the rest of the world as if it were `the only way' \cite{mohamedDecolonial2020, varshneyDecolonial2023}.

\section{Conclusion}
Practitioners of LLM alignment should avoid relying on empty signifiers, and instead, be more precise in what they are attempting to achieve through empirical alignment datasets. In this paper, we presented a framework for communicating which dimensions are measured as alignment signals and how these dimensions are defined. Without such a shared language, there are dangers from obscurities: specific local particularities may be disguised or passed off as universalities, enforcing primarily western and narrow meanings across peoples, countries or cultures. The discourse around alignment carries with it the ideological imprints of various stakeholders---whether this be technology designers, data labellers, or internet users---and we need to document the role that these humans play in shaping model behaviour. We hope this work encourages such critical reflections and supports transparent communication in alignment efforts that must replace the abstract with the empirical and the rhetorical with the actionable.

\section*{Social Impact Statement}
Our work is intended primarily for practitioners (whether this be industry, government, academia, open-source communities or other model developers). We provide a tangible framework for operationalising abstract notions of alignment into measurable empirical signals. The broader impact of our work is in clarifying, conceptualising and re-communicating the empirical alignment landscape via documentation of intents. We envisage the practical applications as two-fold. First, our framework has impact as a development tool which can be applied \textit{ex-ante}, to initiate or guide the process of building datasets for LLM alignment and to iterate on assumptions or aims along the way. Secondly, our framework has impact as a communication tool, which can be applied \textit{ex-post} to empirical alignment research so that achievements and framings can be reflected upon and clearly described to other members of the community or external stakeholders using a shared vocabulary. It is important to note that we do not advocate for adopting one approach over another, nor suggest that occupying one quadrant is somehow `better'---that is, we make no normative calls on what are the right paradigms to conceptualise ``alignment'' in LLMs, or how decisions ``should'' be made. The provision of our framework is not for predicting how future LLMs will integrate with wider society and its diverse members, or forecasting where critical mass will accumulate in different quadrants. However, we do believe our work clearly adds value to the \textit{societal} alignment of LLMs at a meta-level. Even though we do not specify which direction to move in, we strongly advocate for knowing and reporting your coordinates. This shared language better conditions what to \textit{expect} when we interact with a model that is said to be \textit{aligned}. By reducing the expectations gap, a greater degree of transparency, documentation and mutual understanding is likely to have a net positive effect on the societal impacts of LLMs, irrespective of the precise development decisions made in the near and distant future.

\section*{Acknowledgements}
As a component of a wider research agenda on optimising feedback between human-and-model-in-the-loop, this paper has received funding from the MetaAI Dynabench grant. H.R.K’s PhD is supported by the Economic and Social Research Council grant ES/P000649/1. P.R received funding through the INDOMITA project (CUP number J43C22000990001) and the European Research Council (ERC) under the European Union’s Horizon 2020 research and innovation program (No. 949944, INTEGRATOR). We particularly want to thank Andrew Bean (University of Oxford) who greatly assisted in classifying and coding the relevant literature that populates our framework, and Betty Hou (New York University) who provided valuable feedback and spurred interesting discussion.

\bibliographystyle{acl_natbib}
\bibliography{refs, in_review, out_review}



\end{document}